\pdfoutput=1
\documentclass[11pt]{article}

\usepackage{ACL2023}

\usepackage{times}
\usepackage{latexsym}

\usepackage[T1]{fontenc}
\usepackage[utf8]{inputenc}

\usepackage{microtype}

\usepackage{inconsolata}

\usepackage{dialogue}
\usepackage{graphicx}
\usepackage{booktabs}
\usepackage{multirow}
\usepackage{comment}
\usepackage{subcaption}

\title{``Are you telling me to put glasses on the dog?'' Content-Grounded Annotation of Instruction Clarification Requests in the CoDraw Dataset}

\author{Brielen Madureira$^{\mathbf{1}}$ \hspace{10mm}  David Schlangen$^{\mathbf{1, 2}}$ \\
$^{\mathbf{1}}$Computational Linguistics, Department of Linguistics \\ University of Potsdam, Germany \\
$^{\mathbf{2}}$German Research Center for Artificial Intelligence (DFKI), Berlin, Germany \\
  \texttt{\{madureiralasota,david.schlangen\}@uni-potsdam.de}}

\begin{document}
\maketitle
\begin{abstract}
Instruction Clarification Requests are a mechanism to solve communication problems, which is very functional in instruction-following interactions. Recent work has argued that the CoDraw dataset is a valuable source of naturally occurring iCRs. Beyond identifying when iCRs should be made, dialogue models should also be able to generate them with suitable form and content. In this work, we introduce CoDraw-iCR (v2), extending the existing iCR identifiers with fine-grained information grounded in the underlying dialogue game items and possible actions. Our annotation can serve to model and evaluate repair capabilities of dialogue agents.
\end{abstract}

\section{Introduction}
\label{sec:intro}
\begin{figure}[ht!]
    \centering
    \includegraphics[trim={0 5cm 9.5cm 0},clip,width=0.9\linewidth]{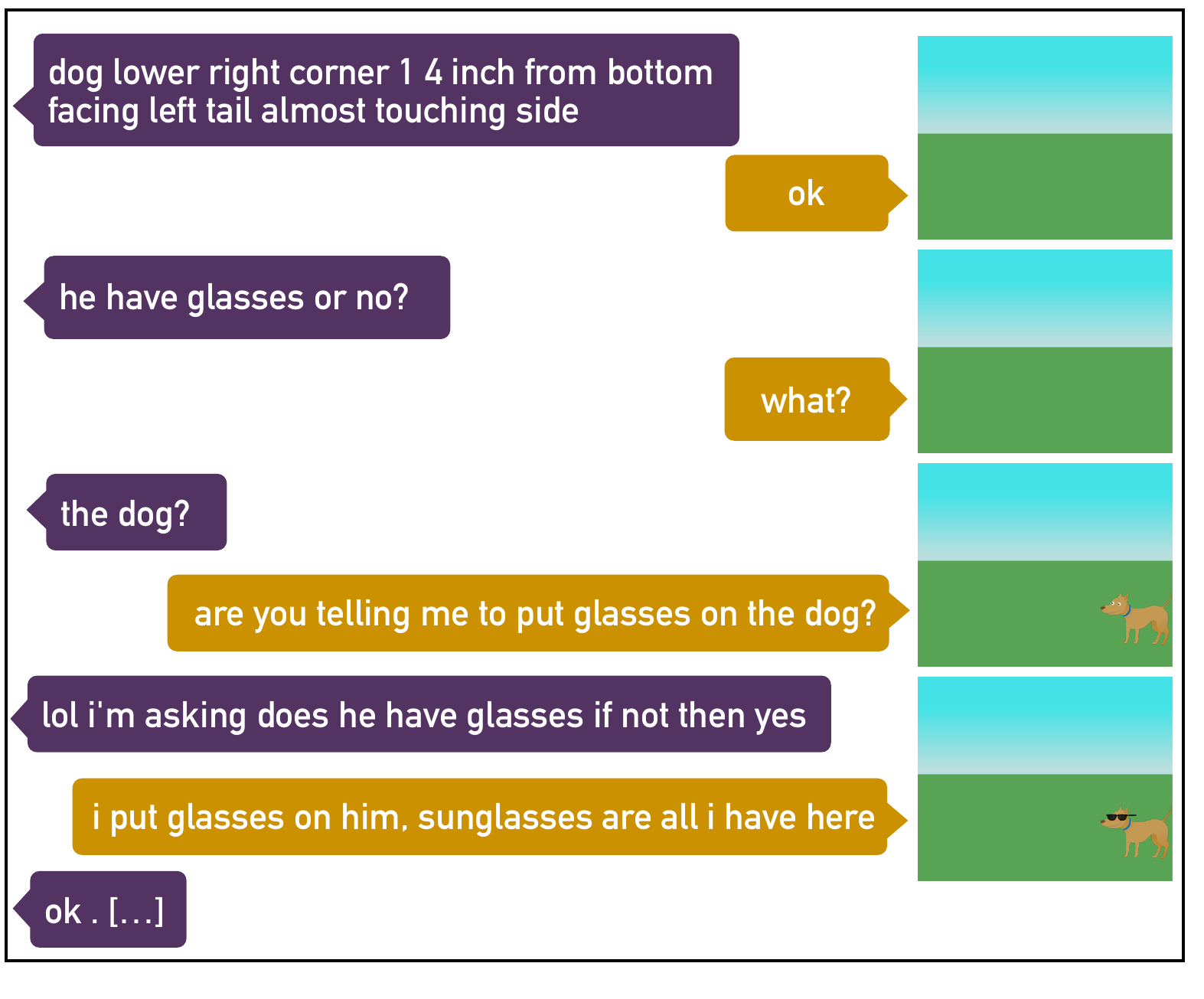}
    \caption{A communication problem occurring and being resolved with the aid of clarification requests in an instruction following interaction (CoDraw, ID 9429, \href{https://creativecommons.org/licenses/by-nc/4.0/}{CC BY-NC 4.0}, scene from \citet{zitnick2013bringing}). When an instruction is not clear enough, the instruction follower asks for clarification, in order to act accordingly (here, placing cliparts in the scene).}
    \label{fig:example-icr}
    \vspace{-0.2cm}
\end{figure}

If someone requests you to put glasses on a dog, you may doubt yourself: \textit{Is that really what I am supposed to do?} Before attempting that, you'd likely seek confirmation, for instance, by posing a clarification request. In real life, dogs do fine without glasses, but, as we see in Figure \ref{fig:example-icr}, that is indeed a correct action in the context of a scene construction dialogue game. 

In instruction following settings, ambiguous or underspecified instructions may elicit clarification requests when the instruction follower realises they cannot act properly without further information. These are Instruction Clarification Requests (iCRs), as defined by \citet{madureira-schlangen-2023-instruction}, namely CRs that occur in Clark's 4th level of communication \citep{clark1996using}, when an instruction is understood generally, but not at the level of uptake \citep{schloder-fernandez-2014-clarification}.

We have recently argued that the CoDraw dataset \citep{kim-etal-2019-codraw} is a rich and large source of spontaneous iCRs \citep{madureira-schlangen-2023-instruction}. We identified iCRs among all instruction follower utterances and proposed using the annotation to model the tasks of knowing \textit{when} to ask and to reply to an iCR. However, knowing \textit{what} and \textit{how} to ask are also topical devices for a competent instruction follower dialogue model. To account for that, we continue this initiative by adding information about the \textit{content} and \textit{form} of iCRs, in order to allow modelling and evaluating the subsequent task of \textit{generating} iCRs, not yet explored in this corpus. 

In this work, we describe our annotation procedure and present a corpus analysis of CoDraw-iCR (v2). Our annotation complements CoDraw-iCR (v1) by adding mood categories and by mapping each utterance to its corresponding objects and action-related attributes.  We show that this sample is an appealing ensemble of mostly unique surface forms through which interesting relations in co-occurring objects and attributes emerge, making it a handy resource for further CR research. We find that iCRs are mostly posed right after an unclear instruction and typically trigger a response at the next turn, so dialogue models should also know how to react timely. The data is available at \url{https://osf.io/gcjhz/}.

% Adding here to make sure it appears in the right page
\begin{table*}[ht!]
\centering
\hspace*{-1cm}
   
    { \setlength{\tabcolsep}{3pt}
    \small
    \begin{tabular}{r r p{4.3cm} p{8.5cm}}
    \toprule
        &   \textbf{Category} & \textbf{Description} & \textbf{Values} \\
    \cmidrule(r){2-4}
        \textbf{Form} & Mood & the iCR surface form & \texttt{declarative, polar question, alternative question, wh- question, imperative, other} \\
        
    \cmidrule(r){2-4}
        \textbf{Locale} & Source & relation to preceding IG utterance & bool, \texttt{1} if iCR is about instruction in immediately preceding utterance \\
        
                        & Response & relation to following IG utterance & bool, \texttt{1} if iCR gets response in immediately following utterance \\
                        
    \cmidrule(r){2-4}
        \textbf{Content} & Quantity & how many objects are mentioned & \texttt{one}, \texttt{two}, \texttt{many}, \texttt{unknown} \\
        
                        & Objects & which objects are mentioned &  clipart identifiers (up to five) \\

                        & Attributes & position & bool, \texttt{1} if iCR is about an object's position in the scene \\

                        & & size  & bool, \texttt{1} if iCR is about an object's size \\

                        & & direction & bool, \texttt{1} if iCR is about an object's direction/orientation \\

                        & & relation & bool, \texttt{1} if iCR is about relations between objects \\

                        & & object disambiguation & bool, \texttt{1} if iCR is disambiguating objects \\

                        & & person disambiguation & bool, \texttt{1} if iCR is disambiguating person's pose or facial expression\\

    \bottomrule
    \end{tabular}
    }

\caption{Content-grounded schema used to annotate iCRs in the CoDraw dataset.}
\label{tab:annotation}
\end{table*}

\section{Related Work}
\label{sec:rel-work}
Clarification Requests are a multifaceted phenomenon in dialogue, with vast literature on categorising, documenting and modelling their various realisations as well as their relations to other utterances and to the context. Annotation efforts have been conducted to identify their causes \citep{gabsdil2003clarification,rodriguez2004form,bohus-rudnicky-2005-sorry,benotti-2009-clarification,koulouri-lauria-2009-exploring,} forms \citep{purver2003means,rodriguez2004form,rieser-moore-2005-implications,deits-2013-clarifying,khalid-etal-2020-combining,gervits-etal-2021-agents} and readings \citep{purver2003means,gabsdil2003clarification,rodriguez2004form,bohus-rudnicky-2005-sorry,rieser-moore-2005-implications,rieser-etal-2005-corpus,kato2013clarifications,liu-etal-2014-detecting,braslavski2017cqa,benotti-blackburn-2021-recipe,shi-etal-2022-learning}.

Still, it remains an open research area; in particular, we cannot delineate yet to what extent CR mechanisms can be learnt via data-driven methods \citep{benotti-blackburn-2021-recipe}, and dealing with underspecifications is still hard for pretrained language models \citep{li2022asking}.

\citet{benotti-blackburn-2021-recipe} have recently raised awareness to the different world modalities upon which clarifications can be \textit{grounded}, like vision, movement or physical objects. Still, few works exist that systematically map the content of CRs to elements related to the context where they occur \citep{gervits-etal-2021-agents}. Some examples are \citet{benotti2017modeling}, who use a methodology to classify CRs according to why they make implicated premises explicit (\textit{e.g.}~wrong plan, not explainable plan or ambiguous plan in instruction giving), in a corpus that is further analysed in \citet{benotti-blackburn-2021-recipe} with a recipe to detecting \textit{grounded clarifications}. \citet{gervits-etal-2021-agents} propose a fine-grained annotation schema for CR types related to the environment (object location, feature, action, description, etc). The small size of these corpora, however, does not meet the needs of current data-driven methods.

CoDraw \citep{kim-etal-2019-codraw} is a dialogue game where an instruction giver, who sees a clipart scene \citep{zitnick2013bringing}, provides written, turn-based instructions to a drawer, who needs to reconstruct the scene. The drawer sees a gallery of objects (a subset of 58 cliparts) and can place, remove, resize or flip the objects and can also ask questions when they wish. \citet{madureira-schlangen-2023-instruction} have put forward a desiderata for Instruction Clarification Requests datasets suitable for data-driven research and demonstrated that the CoDraw-iCR (v1) dataset meets most requirements: Naturalness, specificity, frequency, diversity and relevance (regularity, according to the authors, requires further investigation). Its size (9.9k dialogues with more than 8k iCRs) also makes it more suitable for neural network-based models.

\section{Fine-Grained Annotation of iCRs}
\label{sec:annotation}
Relying on the existing iCR identification in CoDraw-iCR (v1), we hereby introduce CoDraw-iCR (v2). It contains a fine-grained annotation of the subset of iCR utterances in CoDraw, using categories that are expected to be directly relevant for generation (form and content), as summarised in Table \ref{tab:annotation}. 

\paragraph{Motivation} In CoDraw-iCR (v1), the annotators identified, with high agreement, an emblematic dialogue act, whose cause and function are demarcated well: iCRs that happen on Clark's 4th level of communication, their \textit{source utterances} (\textit{i.e.}~the utterance where the communication problem originates) occur during instruction giving, and their purpose is getting an appropriate \textit{response} that enables them to decide how to follow the instruction by making actions. Their realisation, however, evinces many other degrees of variation. An adequate instruction follower model playing the CoDraw dialogue game has a series of decision-making steps to perform when it comes to iCRs. It must detect the time to ask, decide what to ask about in terms of game objects and actions, and define the surface form to realise the iCR. The existing annotation can directly inform the training process for the first, but the other two are left at the mercy of the capabilities of end-to-end models to learn them implicitly from the data. Alternatively, our more detailed annotation can guide the learning process more explicitly and also enrich the evaluation of generated iCRs. 

\begin{figure}[ht!]
    \centering
    \includegraphics[trim={0cm 0cm 0cm 0cm},clip,width=\linewidth]{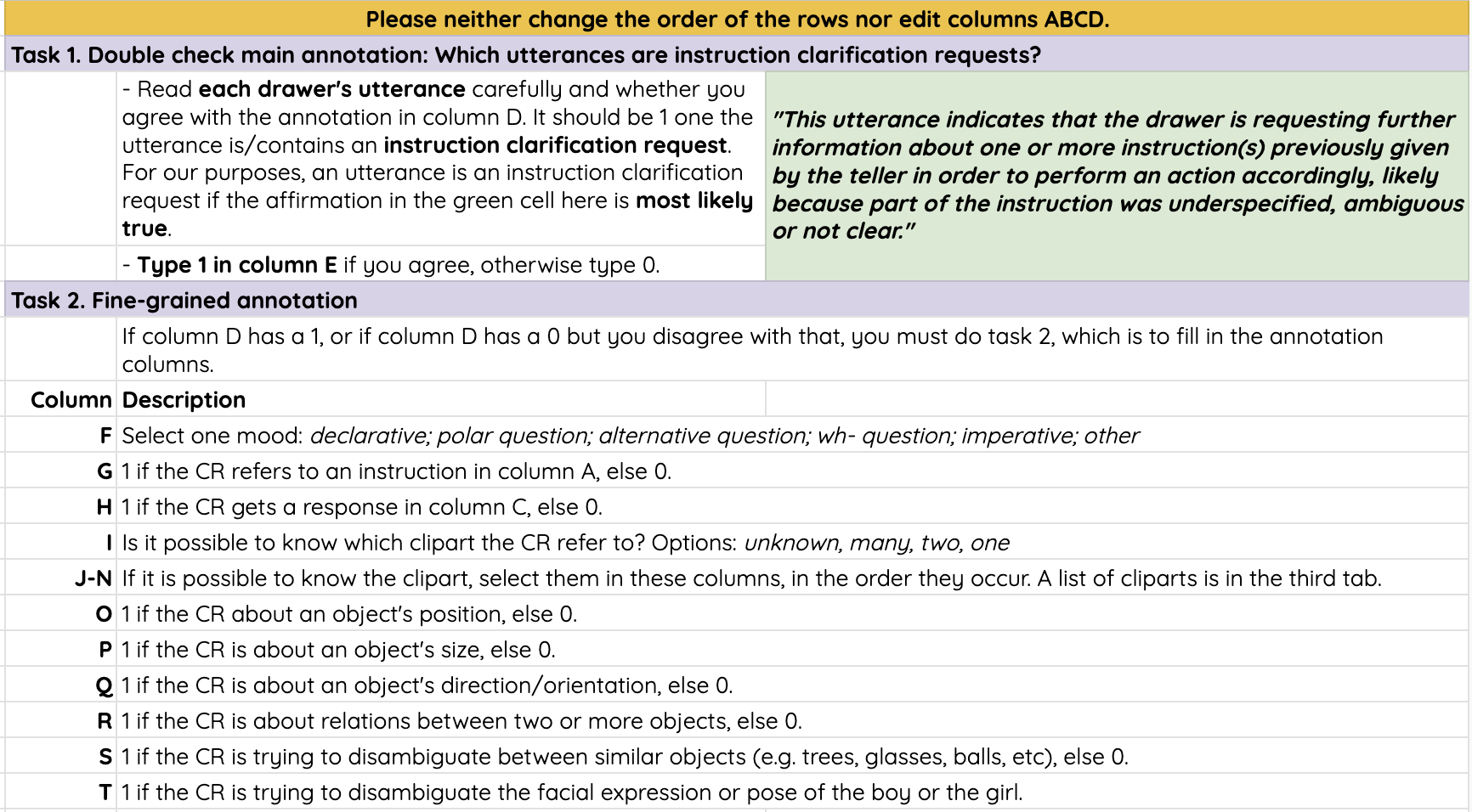}
    \caption{Instructions provided to the annotator.}
    \label{fig:instructions}
\end{figure}

\begin{figure}[ht!]
    \centering
    \includegraphics[trim={1.5cm 2cm 3.5cm 1cm},clip,width=0.95\linewidth]{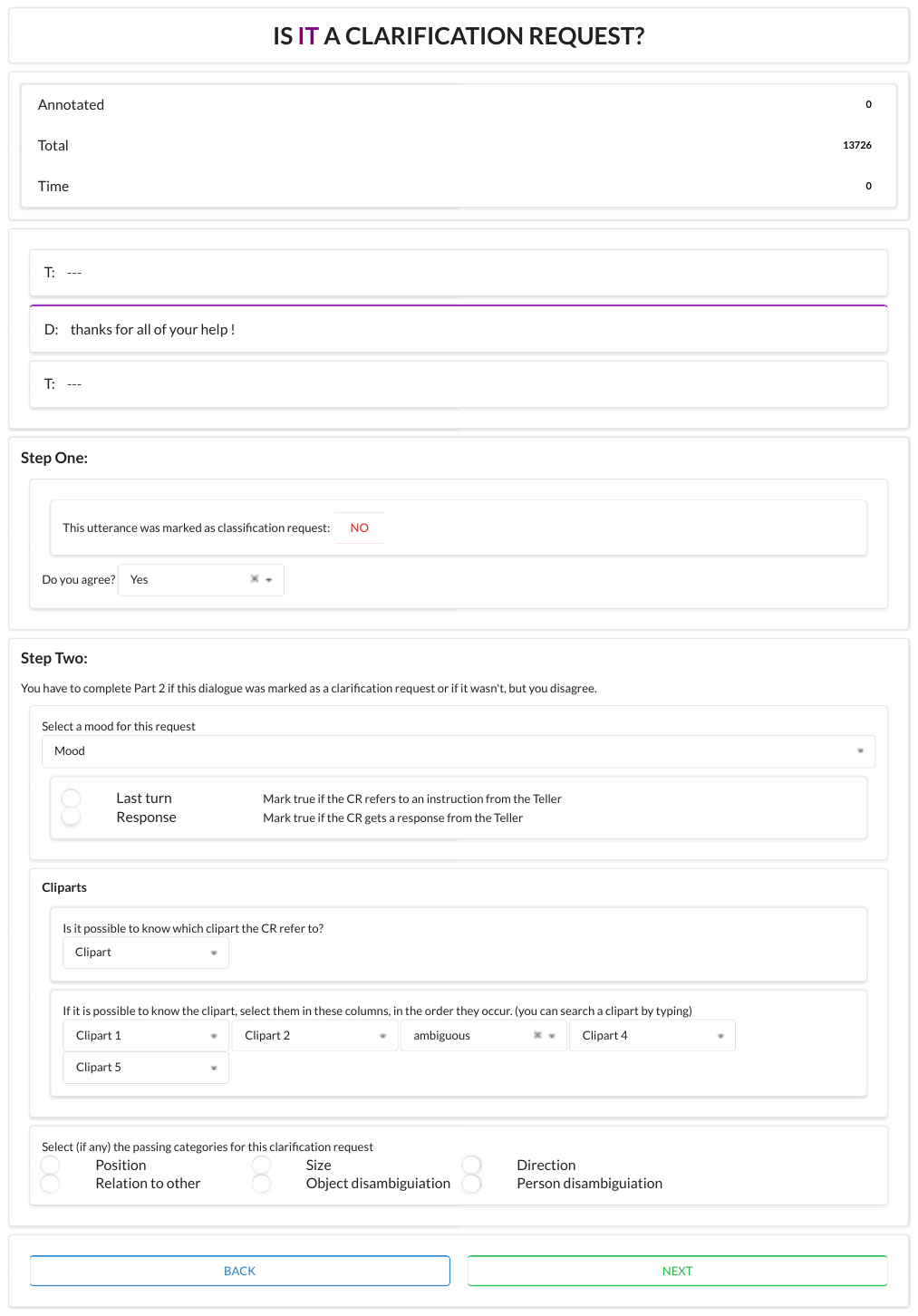}
    \caption{The GUI with the iCR annotation schema.}
    \label{fig:gui}
\end{figure}

\paragraph{Procedure} The annotation was performed by a fluent non-native English speaker working as a student assistant at our lab and paid according to the national regulations. The person is a female computational linguistics bachelor student who went through a learning phase to familiarise herself with the CR literature and with CoDraw objects and rules. The annotation instructions were given as shown in Figure \ref{fig:instructions}. Utterances were then presented in an internally developed graphical user interface where all the predefined values were available (see Figure \ref{fig:gui}, Step Two). The immediately previous and next turns by the instruction giver were shown as context. Some less evident cases were discussed with the author; the main decisions are documented in the data repository. The annotation was performed in a period of around 4 months. 

\paragraph{Locale} Given that iCRs are typically a local phenomenon \citep{rodriguez2004form,schlangen-fernandez-2007-beyond,rieser-moore-2005-implications}, and also the incremental nature of the task and the turn-based setting, only one utterance before and one utterance after the iCR were presented as context. To validate that assumption, the annotator decided whether the previous utterance is the iCR's \textit{source} utterance and whether the next utterance is or contains a \textit{response}.

\paragraph{Form} We follow the schema proposed by \citet{rodriguez2004form} and annotate the \textit{mood} of the iCRs. For each iCR utterance, the annotator could select among \texttt{declarative}, \texttt{polar question}, \texttt{alternative question}, \texttt{wh- question}, \texttt{imperative} and \texttt{other}. Utterances expressing more than one mood were annotated with all suitable categories.\footnote{Due to a limitation of the GUI script, the exact order could not be preserved.}

% Adding here to make sure it appears in the right page
\begin{figure*}[ht!]
    \centering
    \begin{subfigure}[t]{0.3\textwidth}
        \centering
        \includegraphics[trim={0.3cm 0cm 0cm 0},clip,height=3.7cm]{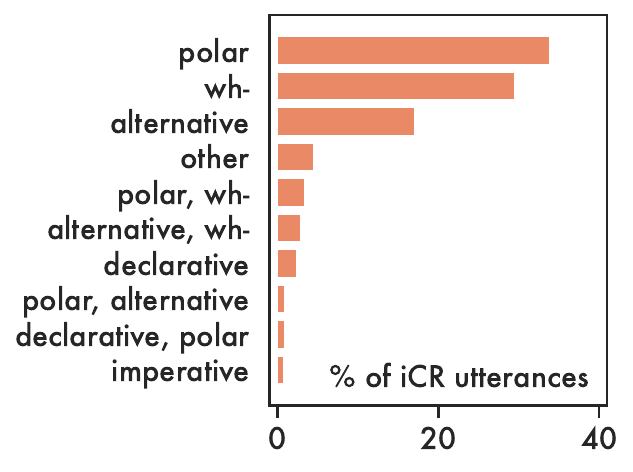}
        \caption{Ten most common moods}
        \label{fig:mood-dist}
    \end{subfigure}%
    ~ 
    \begin{subfigure}[t]{0.4\textwidth}
        \centering
        \includegraphics[height=3.7cm]{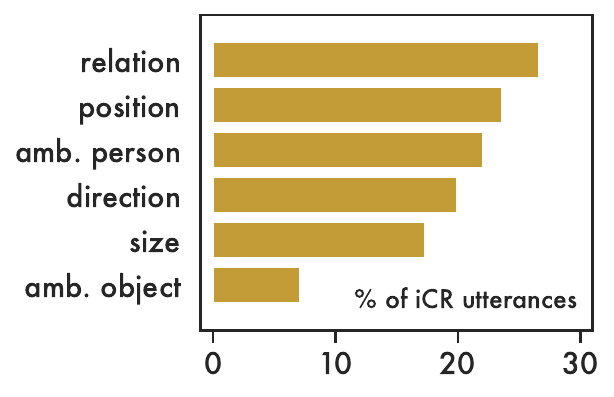}
        \caption{Relative frequency of attributes}
        \label{fig:attr-dist}
    \end{subfigure}%
    ~ 
    \begin{subfigure}[t]{0.3\textwidth}
        \centering
        \includegraphics[height=3.7cm]{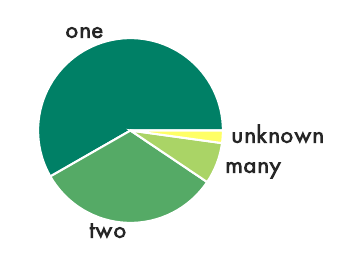}
        \caption{Quantity of mentioned cliparts}
        \label{fig:density}
    \end{subfigure}

    \vspace{1.5cm}

    \begin{subfigure}[t]{0.3\textwidth}
        \centering
        \includegraphics[height=4.2cm]{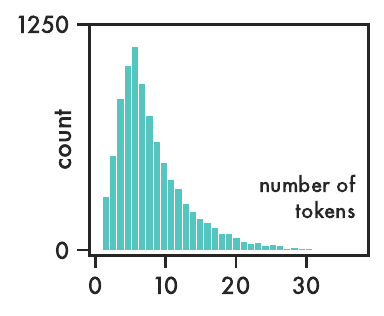}
        \caption{Distribution of length}
        \label{fig:icr-len}
    \end{subfigure}%
    ~ 
    \begin{subfigure}[t]{0.4\textwidth}
        \centering
        \includegraphics[trim={0cm 0.4cm 0cm 0cm},clip,height=4.2cm]{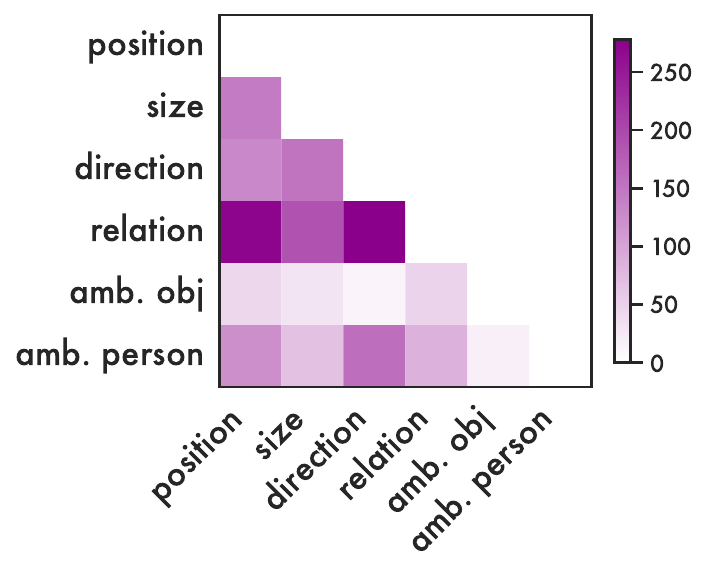}
        \caption{Frequency of co-occurring attributes.}
        \label{fig:attr-inter}
    \end{subfigure}%
    ~ 
    \begin{subfigure}[t]{0.3\textwidth}
        \centering
        \includegraphics[height=4.2cm]{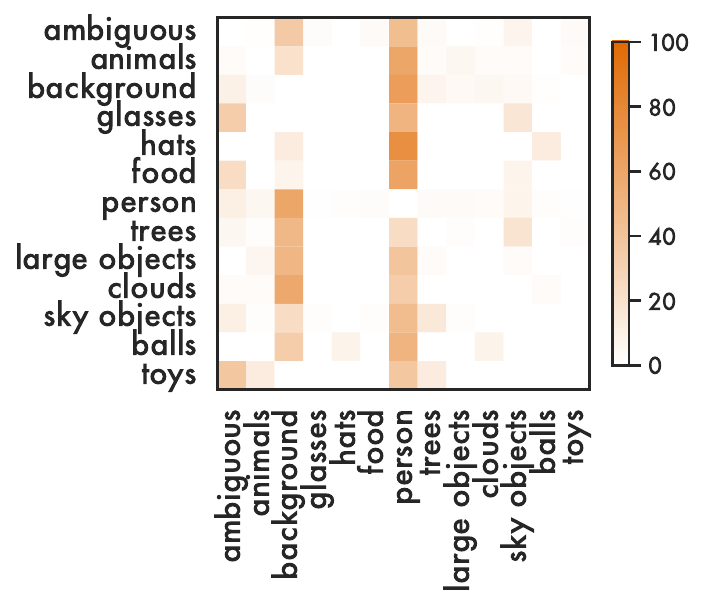}
        \caption{Co-occurrence of thematic object categories, distribution by row in \%.}
        \label{fig:cat-cooc}
    \end{subfigure}
    
    \caption{Overview of the distributions of annotated categories in CoDraw iCR utterances.}
\end{figure*}

\paragraph{Content} The main goal of our fine-grained annotation is to have categories grounded in the underlying dialogue game. It is similar to the types in \citet{gervits-etal-2021-agents}, but adapted to CoDraw, where two aspects are relevant: Objects and actions. Objects are the cliparts, and actions refer to the possible manipulations of their attributes: Size, position, direction and their relation to the scene and to other objects. Upon a close qualitative examination of some dialogues, we also identified iCRs being used to disambiguate between related cliparts, either because multiple types exist (\textit{e.g.}~trees and hats) or, in case of the boy and girl, their pose and facial expression. Therefore, the content annotation was twofold. First, the annotator decided whether it is possible to identify the cliparts being mentioned in the utterance. They were grouped in an aggregate category of quantity (\texttt{one}, \texttt{two}, \texttt{many} or \texttt{unknown}) and then listed (up to five) in the order they occur. Besides the game cliparts, we also added a category for the background (which is the same for all scenes). Following a similar approach in \citet{rodriguez2004form}, we introduced six \textit{ambiguity classes} to account for cases when it was not possible to precisely detect the clipart, five for specific groups of related objects (tree, ball, cloud, hat, glasses) and one for other ambiguities. Then, non-mutually exclusive binary labels were assigned if the iCR was about an object's position in the scene, size and direction, its relation to other objects, and disambiguation of object or person.

\section{Corpus Analysis}
\label{sec:analysis}
In this section, we conduct a detailed examination of the annotated utterances, among which 8,765 utterances (7,710 types) were identified as iCRs.\footnote{The absolute numbers differ slightly from \citet{madureira-schlangen-2023-instruction} due to the inclusion of the in-context annotation of duplicate types.} We present a descriptive analysis enriched with examples of each annotation category, anchoring them in the linguistic components of the game. 

\subsection{Locale}

    The immediately preceding instruction giver utterance is the source utterance (\textit{i.e.}~the utterance where the communication problem manifests) for 80.26\% of the iCR utterances, similar to what is reported by \citet{purver2003means} but around 15\% less than in the corpus study by \citet{rodriguez2004form}. 78.49\% of the iCR utterances get a response from the instruction giver in the immediately following turn. For 63.85\% of them, both conditions are true. This corroborates the assumption that iCRs are usually a local phenomenon in CoDraw, so the context we use is enough for our purposes of annotating form and content.

\begin{table}[ht!]
\centering
   
    { \setlength{\tabcolsep}{6pt}
    \small
    \begin{tabular}{r p{5cm}}
    \toprule
    \textbf{polar}          & is girl angry? \\
                   & large cloud is left? \\
                   & so half sun is visible? \\
    \textbf{wh }            & what are they doing? \\
                   & which tree and what size? \\
                   & how are the boys arms? \\
    \textbf{alternative}    & is she large or small \\
                   & is the balloon in the air or sitting on the ground? \\
                   & right or left or center \\
    \textbf{declarative}    & i don't see a baseball available. \\
                   & you said the tree is on the right side \\
                   & i don't understand where the basketball is supposed to be \\
    \textbf{imperative }    & describe the boy please. \\
                   & confirm the boy is about a half inch from the left of the scene. \\
                   & please clarify \\
    \bottomrule
    \end{tabular}
    }

\caption{Example utterances for each type of mood in CoDraw iCRs.}
\label{tab:mood-examples}
\end{table}

\subsection{Form}

    Examples for each mood are shown in Table \ref{tab:mood-examples}. The average number of tokens in iCR utterances is 8.39, around four times the average length of all instruction follower's utterances, which contained a large portion of very short acknowledgements like \textit{ok}. The distribution is illustrated in Figure \ref{fig:icr-len}. They are realised in many surface forms, ranging from short and generic (\textit{sorry?}), to very specific (\textit{owl is med?}), to long and verbose (\textit{is the girl sitting or standing i need to know as there are multiple options and her expression as well}). Figure \ref{fig:mood-dist} shows the relative frequency of the ten most common moods. Polar questions are the most common, followed by wh-questions and alternative questions. Declarative and imperative moods are much less frequent. In 11.5\% of the cases, the instruction follower uses more than one mood, \textit{e.g.}~by asking more than one CR in a turn, or even integrating them in the same sentence. Examples with multiple moods are \textit{which way is the bolt and is it touching the ground or above it?} (asking about two different attributes with a wh-question and an alternative question) and \textit{which tree? apple or bushy tree?} (refining the first wh-question to make the CR more specific with an alternative).

\subsection{Content}

    We now turn to analyse the two aspects of iCR content, namely objects and manipulable attributes, which directly map to the game objects and actions. As we will see, the iCRs cover all available objects and are well distributed among actions.

    \begin{table}[ht!]
\centering
   
    { \setlength{\tabcolsep}{6pt}
    \small
    \begin{tabular}{r p{5cm}}
    \toprule
    \textbf{position}                  &  campfire on right or left edge? \\
                                       &  tree top and left edge cut? \\
                                       & so the girl is in the middle of the scene? \\
    \textbf{relation}                  & left handle touching sun? \\
                                       & where is the soda located on the table? \\
                                       & which hands are they holding things? \\
    \textbf{direction}                 & facing left or right? \\
                                       & which direction is rocket facing? \\
                                       & cat facing left or right \\
    \textbf{size}                      & ok and girl size \\
                                       & sun big? \\
                                       & are you sure the tree is large? \\    
    \textbf{amb. person}          & does he look happy or surprised? \\
                                       & happy or sad girl? \\
                                       & is she standing? \\
    \textbf{amb. object}          & is the tree pointed at the top? \\
                                       & what is the color of hat? \\
                                       & does the cloud have rain or lightning \\

    \bottomrule
    \end{tabular}
    }

\caption{Example utterances for each type of attribute in CoDraw iCRs.}
\label{tab:attr-examples}
\end{table}

\begin{figure}[ht!]
        \centering
        \includegraphics[trim={0cm 0cm 0cm 0cm},clip,width=0.9\linewidth]{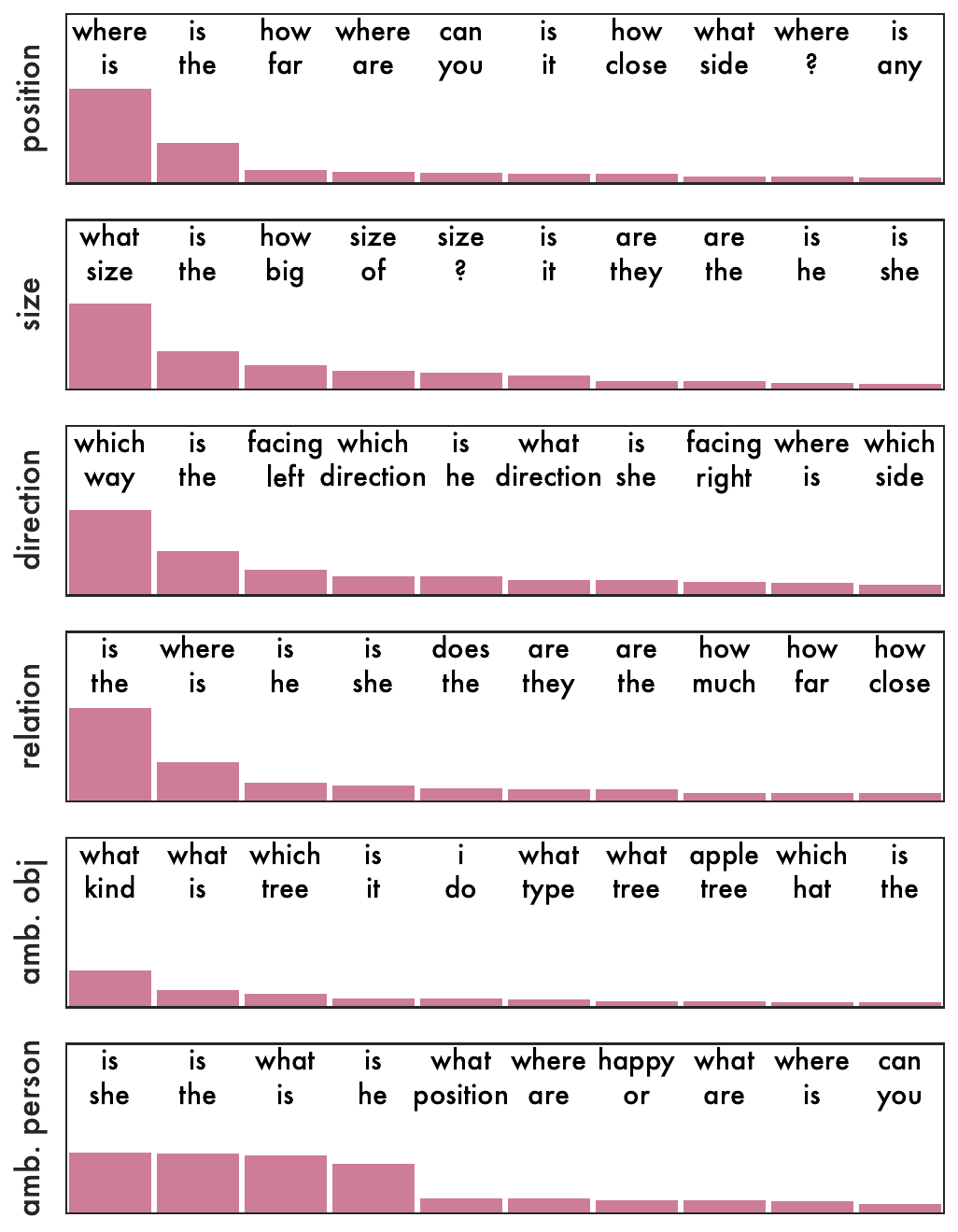}
        \caption{Ten most common initial bigrams for each iCR attribute category.}
        \label{fig:attr-init-bigram}
    \end{figure}

    \paragraph{Attributes} Table \ref{tab:attr-examples} shows examples of utterances seeking to clarify each attribute. In Figure \ref{fig:attr-dist}, we see the relative frequency of each attribute in CoDraw iCRs. While object disambiguation is somewhat less frequent, all other types occur more evenly, each in around 20\% of the iCRs. Attributes are not mutually exclusive: While 82.87\% of the iCR utterances refer to only one attribute, 14.15\% mentions two. Three and four attributes occur together in less than 2\% of the iCRs and are interesting cases of very detailed clarification requests. For example, the utterance \textit{is she sitting, and is she in the sandbox or on the right of it outside?} is about position, relation and ambiguous person and \textit{how close to the bottom are their feet? i have the crown and the boy's hands over grass line. are they smaller?} is about position, size and relation. The frequency of attribute co-occurrence is depicted in Figure \ref{fig:attr-inter}. Position and direction occur very often with relation. Disambiguations of objects occur more rarely with other attributes. We observe some patterns in the ways iCRs are realised for each attribute. The ten most common initial bigrams are shown in Figure \ref{fig:attr-init-bigram}.\footnote{Here, initial \textit{ok} or \textit{ok,} tokens are excluded.} Some bigrams, especially the first ones, look very predictable given the attribute (for instance, keywords like \textit{where}, \textit{close} and \textit{far} are relevant for positioning and \textit{way}, \textit{facing} and \textit{direction} are indicative of direction). On the other hand, \textit{is the} is a versatile initial bigram which is frequent for all attributes. This is pertinent information to be integrated during generation, so that the iCRs sound natural and purposeful. 

    \begin{figure*}[ht!]
        \centering
        \includegraphics[trim={0cm 0cm 0cm 0cm},clip,width=\linewidth]{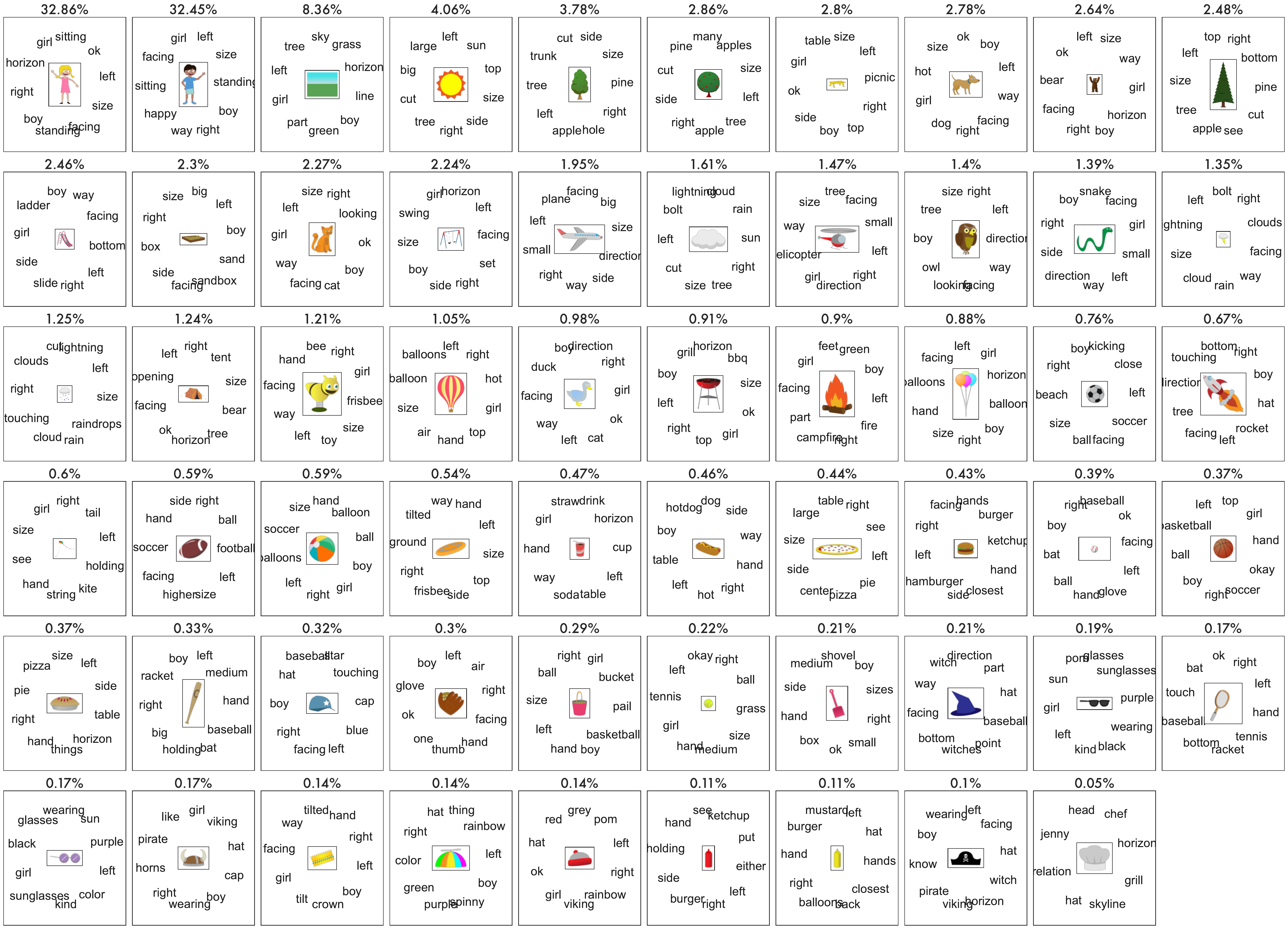}
        \caption{The ten most common tokens (excluding stopwords) associated with each game object in iCRs. The percentages are the relative frequency over all iCR utterances. Clipart images from \citet{zitnick2013bringing}.}
        \label{fig:clip-vocab}
    \end{figure*}

    \begin{figure*}[ht!]
        \centering
        \includegraphics[trim={0cm 0.4cm 0cm 0cm},clip,width=\linewidth]{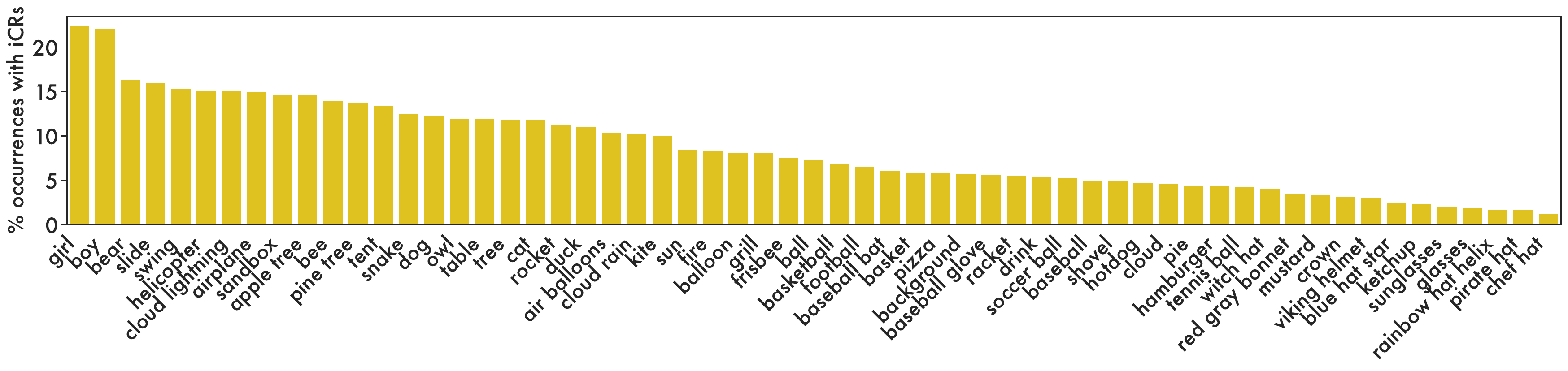}
        \caption{Dialogues with at least one iCR about each clipart, normalised by their frequency across scenes.}
        \label{fig:perc-clips}
    \end{figure*}
    
\begin{figure*}[ht!]
        \centering
        \includegraphics[trim={0cm 0cm 0cm 0cm},clip,width=\linewidth]{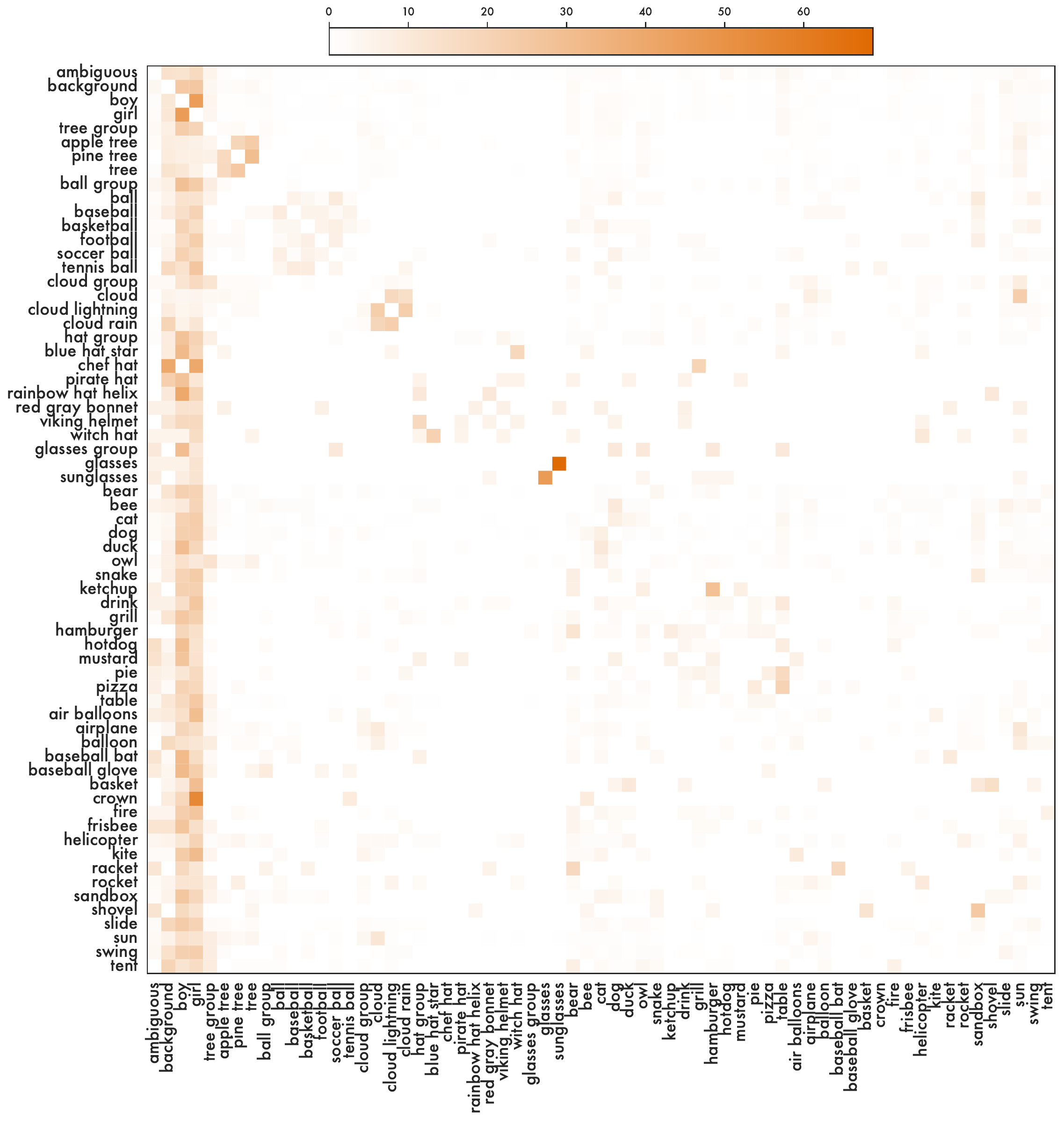}
        \caption{Co-occurrence of objects, distribution by row in \%.}
        \label{fig:clip-cooc}
    \end{figure*}

    \paragraph{Objects} When it comes to the game objects, the relevant aspects are \textit{quantity} (how many objects are mentioned in an iCR), \textit{frequency} (which objects lead to more need to clarify) and \textit{co-occurrences} (how the relations between specific objects need to be clarified). In terms of quantity, in Figure \ref{fig:density} we see that 58.28\% of the iCRs refer to only one object, and a considerable portion refers to two (32.29\%). The relative frequency of all 58 game objects (and the background) in the set of iCRs and the most common vocabulary associated to each of them is shown in Figure \ref{fig:clip-vocab}. The person cliparts are the most common.\footnote{For illustration purposes, only one pose for the boy and the girl is shown in Figure \ref{fig:clip-vocab}.} Next comes the background, whose horizon is commonly used as a point of reference when positioning objects. The three types of trees are also a common subject. Multiplicity is, however, not an explaining factor on its own, because the different hats and glasses appear much less often. Objects frequency in iCRs is actually positively correlated with their frequency in the scenes (Spearman $\rho=0.82$). To explore it in more detail, Figure \ref{fig:perc-clips} shows, for each object, the percentage of times they triggered at least one iCR in a dialogue among all scenes in which they appeared. Higher percentages mean that the object is frequently involved in communication problems when it is in a scene. The boy and the girl, each of them occurring in more than 90\% of the scenes, very often require clarification. This is likely due to their various poses and facial expressions that require disambiguation and also to their relation to other cliparts (for instance, hats on their heads or their hands holding objects). The thematic category of large objects and some objects in multiple forms (cloud and tree) also bring on many iCRs. However, some ambiguous objects, like chef hat and pirate hat or glasses and sunglasses, very rarely trigger clarification.\footnote{Further investigation is necessary to understand if they indeed trigger precise referring expressions. There may also be an effect of the subset of available cliparts in the gallery. Although many hats exist in the game, if the gallery shown to the instruction follower only contains one type, no disambiguation will be needed even in face of an underspecified instruction.} Figure \ref{fig:clip-cooc} shows the co-occurrences of cliparts, with distributions by row. We can see clusters in ambiguous objects like balls and clouds, but semantically related objects like shovel and sandbox, racket and baseball bat, and table and pizza also tend to co-occur. Persons and the background occur often with almost all other objects. In Figure \ref{fig:cat-cooc}, objects are grouped into thematic categories. Besides the evident ubiquity of person and background, some less obvious relations become clear: hats and balls, sky objects and glasses, sky objects and trees, balls and clouds.

\subsection{Interrelations}

\begin{figure}[ht!]
        \centering
        \includegraphics[trim={0cm 0cm 0cm 0cm},clip,width=\linewidth]{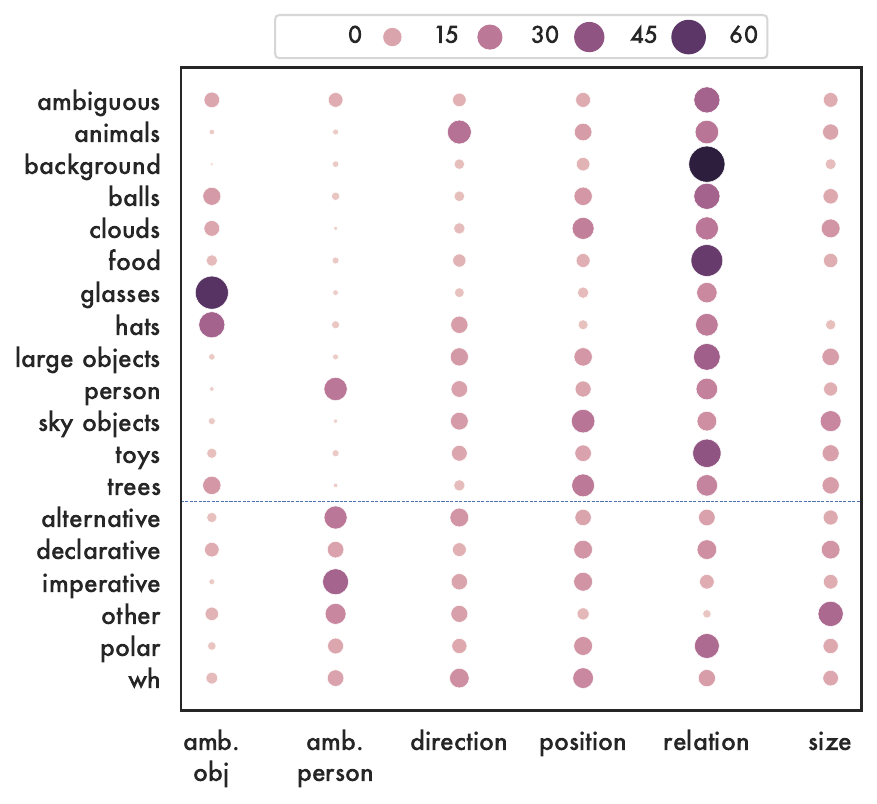}
        \caption{Co-occurrence of clipart categories and mood with attributes, distribution by row in \%.}
        \label{fig:clip-attr-mood}
    \end{figure}

    The distribution of attribute for each object category and mood is shown in \ref{fig:clip-attr-mood}. Polar questions are often used to clarify relations, while alternative and imperative are common to disambiguate persons. Wh-questions and declarative are used more uniformly for all attributes. When grouped by thematic category, we see that relation is a predominant topic for almost all groups (note, however, that relations always include more than one clipart in the count). Glasses and hats very often require object disambiguation. For sky objects and trees, position is also a usual topic. 

\section{Conclusion}
\label{sec:conclusion}
We have analysed the manifestation of Instruction Clarification Requests in the CoDraw dataset in terms of form (mood and length) and content (objects, attributes and their co-occurrences) and some of their interrelations. Our findings further support the argument that the CoDraw data collection setting was an effective means to elicit iCRs. Even in its controlled environment with a limited number of actions and objects, the resulting iCR utterances are very diverse in their surface form and very fertile in their content. With the release of the annotated data, the community gains a larger resource with sequential, spontaneous iCRs in turn-based dialogues. We aim to encourage more research on modelling clarification requests in instruction following interactions, and also to enable more detailed evaluation of iCR generation.

\paragraph{Limitations} Given the need for large scale corpora for data-driven methods, trading some of the ecological validity in the annotation process for machine-learnability was necessary. Due to the availability of resources for this project, the annotation was performed by only one annotator, thus inter-annotator agreement could not be measured. Still, the quality of a sample was verified by the author to during the initial annotation phase. The scenes were not available during annotation, so clipart annotation can contain misunderstandings. The macro categories for ambiguous cases help alleviate that, but we still observed that misclassifications occur in some cases. The annotator worked on utterance level but, as we showed, some utterances contain non-iCR content or multiple categories. A possible enhancement is to further segment utterances in order to allow each category to be mapped to sentences or phrases (for mood and attribute) and to tokens or phrases (for objects and their referring expressions), and also to identify the tokens that are unrelated to the current iCR. For the portion of the iCRs whose source utterances occur further back in the dialogue context, or whose responses are not given in the next turn, we currently lack annotation that would allow full examination of their context.

\section*{Acknowledgements}
We are very thankful to our student assistants, Sebastiano Gigliobianco, for performing the initial identification and implementing the GUI and Sophia Rauh, for doing the fine-grained annotation. We also thank the anonymous reviewers for their suggestions on the text.

%\clearpage

\bibliography{anthology,custom}
\bibliographystyle{acl_natbib}

\end{document}